# Unsupervised Activity Discovery and Characterization From Event-Streams *

**Raffay Hamid, Siddhartha Maddi, Amos Johnson, Aaron Bobick, Irfan Essa, Charles Isbell**
College of Computing
Georgia Institute of Technology
Atlanta, GA  30332-0280 USA
{raffay, maddis, amos, afb, irfan, isbell}@cc.gatech.edu

## Abstract

We present a framework to discover and characterize different classes of everyday activities from event-streams. We begin by representing activities as bags of event n-grams. This allows us to analyze the global structural information of activities, using their local event statistics. We demonstrate how maximal cliques in an undirected edge-weighted graph of activities, can be used for activity-class discovery in an unsupervised manner. We show how modeling an activity as a variable length Markov process, can be used to discover recurrent event-motifs to characterize the discovered activity-classes. We present results over extensive data-sets, collected from multiple active environments, to show the competence and generalizability of our proposed framework.

## 1   Introduction

Consider a Loading Dock area with delivery vehicles, people, packages etc. The interaction of these objects would form different events, which in turn constitute the different instances of package-delivery activity. Understanding what is happening in such active settings has many potential applications, ranging from automatic surveillance systems to supporting users in ubiquitous environments. A key step to this end is to discover the kinds of similar activities that frequently occur in a particular domain. Equally important is the question of finding efficient characterizations for these different kinds of activities. In this paper, we tackle the question of activity class discovery and characterization, in the context of analyzing everyday activities.

We begin by introducing a novel representation of activities as bags of discrete event $n$-grams - a perspective different from the previously used grammar driven approaches. This treatment of activities, motivated by some recent developments in natural language processing, allows us to analyze the global structural information of the activities by simply considering its local event statistics.

Based on this activity representation, we formalize the notion of similarity between two activities, taking into account their core structural and event-frequency based differences. We pose activity-class discovery as a graph theoretic problem, and demonstrate how finding *maximal cliques* in edge-weighted graphs can be used to this end.

A concise characterization of these discovered activity-classes is of fundamental interest. Such characterizations can be used for online activity classification and detecting non-regularities in activities. Taking on some of the previous work in the field of bio-informatics, we formalize this problem as finding predictably recurrent event subsequences (*motifs* - defined formally in Section 6.1) using *variable-memory Markov chains*.

The main contributions of this work are:
- A novel representation of activities as bags of discrete event $n$-grams.
- An algorithm for unsupervised discovery of disjunctive activity groups.
- A framework for unsupervised discovery of predictably recurrent event motifs for activity class characterization.

## 2   Previous Work

In the past, various approaches for activity representation have been fundamentally grammar-driven (see *e.g.* [Ivanov and Bobick2000], [Minnen *et al.*2003]). In this work we propose to treat activities as bags of event $n$-grams, which allows us to extract the global structural information of an activity, by simply considering its event statistics at a local scale.

Although the idea of discovering activity-classes has been previously explored in such fields as *network intrusion detection*, it has only recently been applied to everyday activities. Our approach towards this problem is novel in a few key aspects. Unlike [Hongeng and Nevatia2001] which require *a priori* expert knowledge to model the activity-classes in an environment, we propose to discover this information in an *unsupervised* fashion. Since event-monograms, as used in [Zhong *et al.*2004] and [Stauffer and Grimson2000], do not capture the temporal information of an activity, we use higher order event $n$-grams to capture this information more efficiently.

---

*This project was funded by DARPA as a part of the CALO project - grant number: SUBC 03-000214

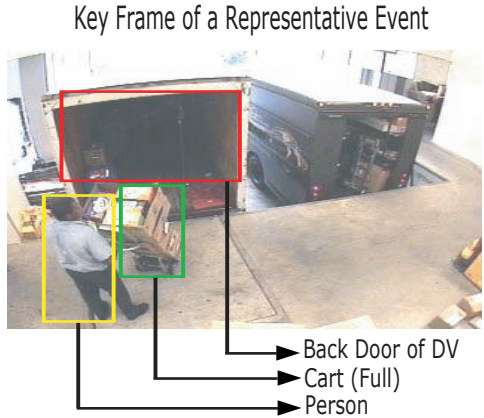

Figure 1: A Person pushes a Cart carrying Packages into the Back Door of a Delivery Vehicle.

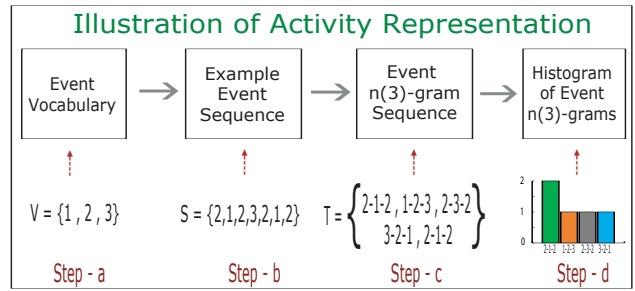

Figure 2: Transformation of an example activity from sequence of discrete events to histogram of event $n$-grams. Here the value of $n$ is shown to be equal to 3. V is event vocabulary, S is event sequence, and T is sequence of overlapping $n$-grams. Step-d shows the non-zero $n$-gram counts of V.

Numerous solutions to the problem of discovering important recurrent motifs in sequential data have been proposed (see *e.g.* [Oates2002] and [Chiu *et al.*2003]). Work done in [Weinberger *et al.*1995] and [Ron *et al.*1996] present techniques for learning variable-memory Markov chains from training data in an unsupervised manner. The variable-memory elements in these Markov chains can be thought of as motifs that have good predictive power of the future events. However they presume the availability of pre-classified data. Moreover, their approach does not filter out the motifs that are common in multiple classes. We modify the work in [Weinberger *et al.*1995] to handle data from multiple classes, finding motifs that are maximally mutually exclusive amongst activity-classes. This forms a nice continuum between the activity-class discovery, and characterization. Moreover, instead of sequentially finding individual motifs and masking them out from the sequences as proposed in [Bailey and Elkan1994], our scheme simultaneously finds all the motifs in the data in one pass. This allows us to find partially overlapping motifs.

## 3 Activity Representation

An active environment consists of animate and inanimate objects interacting with each other. The interaction of these objects in a particular manner constitutes an event. Looking at an activity as a sequence of discrete events, two important quantities emerge:

- *Content* - events that span the activity, and
- *Order* - the arrangement of the set of events.

This treatment of an activity is similar to the representation of a document as a set of words - also known as the Vector Space Model (VSM) [Salton1971], in which a document is represented as a vector of its word-counts, in the space of possible words.

To use such a scheme, we must define a set of possible events (*event vocabulary*) that could take place in the situation under consideration. As this representation is designed to be manipulated by a perceptual system, the events must be chosen such that they are detectable from low-level perceptual primitives. A particular interaction of these perceptual primitives constitute an event. A key-frame of a representative event from one of the active environments that we explored (Loading Dock) is shown in Figure 1.

While VSM captures the content of a sequence in an efficient way, it ignores its order. Because the word content in documents often implies causal structure, this is usually not a significant problem. Generally activities are not fully defined by their event-content alone; rather, they have preferred or typical event-orderings. Therefore a model for capturing the order of events is needed. To this end, we consider histograms of higher order event $n$-grams (see figure 2), where we represent an activity by a sparse vector of counts of overlapping event $n$-grams in a high dimensional space of possible event $n$-grams. Our proposed scheme would capture the activity-structure for domains with substantive structural coherence. It is evident that higher values of $n$ would capture the temporal order information of events more rigidly, and would entail a more discriminative representation.

## 4 Activity Similarity Metric

Sequence comparison is a well-studied problem and has numerous applications [Gusfield1997]. Our view of the similarity between a pair of sequences consists of two factors, the *core structural differences* and differences based on the *frequency of occurrence* of event $n$-grams.

The core structural differences relate to the distinct $n$-grams that occurred in either one of the sequences in a sequence-pair, but not in both. We believe that for such differences, the number of these mutually exclusive $n$-grams is of fundamental interest. On the other hand, if a particular $n$-gram is inclusive in both the sequences, the only discrimination that can be drawn between the sequence pair is purely based on the frequency of the occurrence of that $n$-gram.

Let $A$ and $B$ denote two sequences of events, and let their corresponding histogram of $n$-grams be denoted by $H_A$ and $H_B$. Let $Y$ and $Z$ be the sets of indices of $n$-grams with counts greater than zero in $H_A$ and $H_B$ respectively. Let $\alpha_i$ denote different $n$-grams. $f(\alpha_i|H_A)$ and $f(\alpha_i|H_B)$ denote the counts of $\alpha_i$ in sequences A and B respectively. We define the similarity between two event sequences as:

$$sim(A,B) = 1 - \kappa \sum_{i \in Y,Z} \frac{|f(\alpha_i|H_A) - f(\alpha_i|H_B)|}{f(\alpha_i|H_A) + f(\alpha_i|H_B)} \quad (1)$$

where $\kappa = 1/(||Y|| + ||Z||)$ is the normalizing factor, and $||\cdot||$ computes the cardinality of a set. While our proposed similarity metric conforms to: (1) the property of Identity of indiscernibles, (2) is commutative, and (3) is positive semi-definite, it does not however follow Cauchy-Schwartz inequality, making it a divergence rather than a true distance metric.

# 5 Activity Class Discovery

We assert that the activity-instances occurring in an environment do not span the activity-space uniformly. Rather, there exist disjunctive activity-sets with high internal similarity while low similarity across the sets. This assertion is backed by the assumption that the detected events, constituting activities in an environment, encode the underlying structure of activities [Rosch *et al.*1976].

## 5.1 Activity-Class as Maximal Clique

Starting off with a set of $K$ activity-instances, we consider this activity-set as an undirected edge-weighted graph with $K$ nodes, each node representing a histogram of n-grams of one of the $K$ activity-instances. The weight of an edge is the similarity between a pair of nodes as defined in Section 4. We formalize the problem of discovering activity-classes as searching for edge-weighted maximal cliques[1] in the graph of $K$ activity-instances. Indeed, in the past, some authors have argued that maximal clique is the strictest definition of a cluster [Auguston and Miker1970]. We proceed by finding a maximal clique in the graph, removing that clique from the graph, and repeating this process sequentially with the remaining set of nodes, until there remain no non-trivial[2] maximal cliques in the graph. The leftover nodes after the removal of maximal cliques are dissimilar from most of the nodes.

---

[1]Recall that a subset of nodes of a graph is a clique if all its nodes are mutually adjacent; a *maximal* clique is not contained in any larger clique, a *maximum* clique has largest cardinality.

## 5.2 Maximal Cliques using Dominant Sets

Finding maximal cliques in an edge-weighted undirected graph is a classic graph theoretic problem. Because combinatorially searching for maximal cliques is computationally hard, numerous approximations to the solution of this problem have been proposed (see [Raghavan and Yu1981] and the references within). For our purposes, we adopt the approximate approach of iteratively finding *dominant sets* of maximally similar nodes in a graph (equivalent to finding maximal cliques) [Pavan and Pelillo2003]. Besides providing an efficient approximation to finding maximal cliques, the framework of dominant sets naturally provides a principled measure of the cohesiveness of a class as well as a measure of node participation in its membership class. This measure of class participation may be used for an instance based representation of a clique [Kleinberg1999]. We now give an overview of dominant sets showing how they can be used for our problem.

Let the data to be clustered be represented by an undirected edge-weighted graph with no self-loops $G = (V, E, \vartheta)$ where $V$ is the vertex set $V = \{1, 2, ...K\}$, $E \subseteq V \times V$ is the edge set, and $\vartheta : E \to \mathbb{R}^+$ is the positive weight function. The weight on the edges of the graph are represented by a corresponding $K \times K$ symmetric similarity matrix $A = (a_{ij})$ defined as:

$$a_{ij} = \begin{cases} sim(i,j) & if\ (i,j) \in E \\ 0 & otherwise \end{cases} \quad (2)$$

where $sim$ is computed using our proposed notion of similarity as described in Section4. To quantize the cohesiveness of a node in a cluster, let us define its "average weighted degree". Let $S \subseteq V$ be a non-empty subset of vertices and $i \in S$, such that,

$$awdeg_S(i) = \frac{1}{||S||} \sum_{j \in S} a_{ij} \quad (3)$$

Moreover, for $j \notin S$, we define $\Phi_S$ as:

$$\Phi_S(i,j) = a_{ij} - awdeg_S(i) \quad (4)$$

Intuitively, $\Phi_S(i,j)$ measures the similarity between nodes $j$ and $i$, with respect to the average similarity between node $i$ and its neighbors in $S$. Note that $\Phi_S(i,j)$ can either be positive or negative.

Now let us consider how coupling-weights are assigned to individual nodes. Let $S \subseteq V$ be a non-empty subset of vertices and $i \in S$. The coupling-weight of $i$ w.r.t. $S$ is given as:

$$w_S(i) = \begin{cases} 1 & if\ ||S|| = 1 \\ \sum_{j \in S \setminus \{i\}} \Phi_{S \setminus \{i\}}(j,i) w_{S \setminus \{i\}}(j) & otherwise \end{cases} \quad (5)$$

---

[2]A non-trivial clique has nodes greater than or equal to three.

Moreover, the total coupling-weight of $S$ is defined to be:

$$W(S) = \sum_{i \in S} w_S(i) \qquad (6)$$

Intuitively, $w_S(i)$ gives a measure of the overall similarity between vertex $i$ and the vertices of $S \setminus \{i\}$ with respect to the overall similarity among the vertices in $S \setminus \{i\}$. We are now in a position to define *dominant sets*. A non-empty sub-set of vertices $S \subseteq V$ such that $W(T) > 0$ for any non-empty $T \subseteq S$, is said to be *dominant* if:

1. $w_S(i) > 0, \forall i \in S$, i.e. internal homogeneity
2. $w_{S \cup \{i\}}(i) < 0 \ \forall i \notin S$, i.e. external inhomogeneity.

Because solving Equation 5 combinatorially is infeasible, we use the continuous optimization technique of *Replicator Dynamics* to optimize Equation 5(for details please refer to [Pavan and Pelillo2003]).

## 6 Discovering Motifs in Activity-Classes

Having discovered various activity-classes in an active environment, we now turn our attention to finding interesting recurrent event-motifs in these discovered classes. Some of the previous work done in bio-informatics on finding motifs, presumes the availability of pre-classified data [Bejerano and Yona1999]. Moreover, these approaches do not filter out the motifs that are common in multiple classes. Our proposed scheme discovers activity-classes in an unsupervised manner, and finds patterns that are maximally mutually exclusive amongst activity-classes.

### 6.1 A Definition of Motif

From the perspective of activity discovery and recognition, we are interested in frequently occurring event-sequences that are useful in predicting future events, and can therefore be used for activity class characterization. Following [Weinberger et al.1995], we assume that a class of activity-sequences can be modeled as a variable-memory Markov chain (*VMMC*). We define an event-motif for an activity-class as one of the variable-memory elements of its *VMMC*. We cast the problem of finding the optimal length of the memory element of a *VMMC* as a function optimization problem and propose our objective function in the following.

Let $Y$ be the set of events, $A$ be the set of activity-instances, and $C$ be the set of discovered activity-classes. Let us define a function $\mathcal{U}(a)$ that maps an activity $a \in A$ to its membership class $c \in C$, as explained in Section 5.2. Let us define the set of activities belonging to a particular class $c \in C$ as $A_c = \{a \in A : \mathcal{U}(a) = c\}$. For $a = (y_1, y_2, ..., y_n) \in A$ where $y_1, y_2, ...y_n \in Y$, let $p(c|a)$ denote the probability that activity $a$ belongs to class $c$. Then,

$$p(c|a) = \frac{p(a|c)p(c)}{p(a)} \propto \prod_{i=1}^{n} p(y_i|y_{i-1}, y_{i-2}, ..., y_1, c) \qquad (7)$$

where we have assumed that all activities and classes are equally likely. We approximate Eq 7 by a *VMMC*, $M_c$ to get:

$$\prod_{i=1}^{n} p(y_i|y_{i-1}, y_{i-2}, ..., y_1, c) = \prod_{i=1}^{n} p(y_i|y_{i-1}, y_{i-2}, ..., y_{i-m_i}, c) \qquad (8)$$

where $m_i \leq i - 1 \ \forall \ i$. For any $1 \leq i \leq n$, the sequence $(y_{i-1}, y_{i-2}, ..., y_{i-m_i})$ is called the *context* of $y_i$ in $M_c$( [Weinberger et al.1995]), denoted by $\mathcal{S}_{M_c}(y_i)$. We want to find the sub-sequences which can effeciently characterize a particular class, while having minimal representation in other classes. We therefore define our objective function as:

$$\mathcal{Q}(M_c|A_c) = \gamma - \lambda \qquad (9)$$

where

$$\gamma = \prod_{a \in A_c} p(c|a) \qquad (10)$$

and

$$\lambda = \sum_{c' \in C \setminus \{c\}} \prod_{a \in A_{c'}} p(c'|a) \qquad (11)$$

Intuitively, $\gamma$ represents how well a set of event-motifs can characterize a class in terms of correctly classifying the activities belonging to that class. On the other hand, $\lambda$ denotes to what extent a set of motifs of a class represent activities belonging to other classes. It is clear that maximizing $\gamma$ while minimizing $\lambda$ would result in the optimization of $\mathcal{Q}(M_c|A_c)$. Note that our motif finding algorithm leverages the availability of the discovered activity-classes to find the maximally mutually exclusive motifs. This shows the usefulness of our activity discovery framework as a pre-step to the motif finding scheme.

### 6.2 Objective Function Optimization

We now explain how we optimize our proposed objective function. [Weinberger et al.1995] describe a technique to compare different *VMMC* models that balances the predictive power of a model with its complexity. Let $s$ be a context in $M_c$, where $s = y_{n-1}, y_{n-2}, ..., y_1$, and $y_{n-1}, y_{n-2}, ..., y_1 \in Y$. Let us define the suffix of $s$ as *suffix*$(s) = y_{n-1}, y_{n-1}, ...y_2$. For each $y \in Y$, let $N_{A'}(y, s)$ be the number of occurances of event $y$ in activity-sequences contained in $A' \subseteq A$ where $s$ precedes $y$, and let $N_{A'}(s)$ be the number of occurances of $s$ in activity-sequences in $A'$. We define the function $\Delta_{A'}(s)$ as

$$\Delta_{A'}(s) = \sum_{y \in Y} N(s, y) log\left(\frac{\hat{p}(y|s)}{\hat{p}(y|suffix(s))}\right) \qquad (12)$$

where $\hat{p}(y|s) = N_{A'}(s, y)/N_{A'}(s)$ is the maximum likelihood estimator of $p(y|s)$. Intuitively, $\Delta_{A'}(s)$ represents the number of bits that would be saved if the events following $s$ in $A'$, were encoded using $s$ as a context, versus having

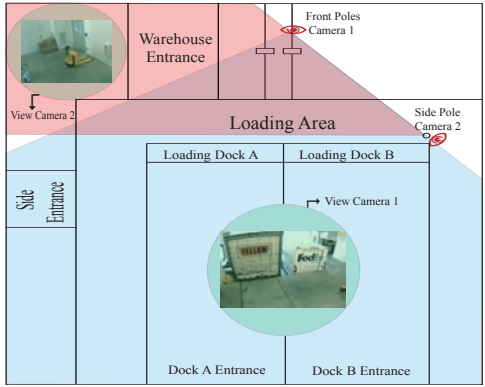
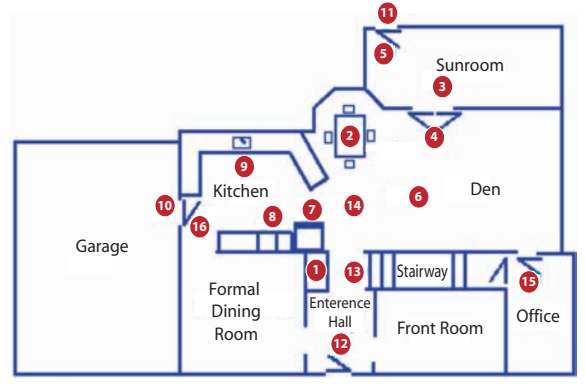

Figure 3: A schematic diagram of the camera setup at the loading dock area with overlapping fields of view (FOV). The FOV of camera 1 is shown in blue while that of camera 2 is in red. The overlapping area of the dock is shown in purple.

Figure 4: A schematic diagram of the strain-gage setup in the house scenario. The red dots represents the positions of the strain gages.

*suffix*($s$) as a context. In other words, it represents how much better the model could predict the events following $s$ by including the last event in $s$ as part of context of these events.

We now define the function $\Psi_c(s)$ (bit gain of $s$) as

$$\Psi_c(s) = \Delta_{A_c}(s) - \sum_{c' \in C \setminus \{c\}} \Delta_{A_{c'}}(s) \qquad (13)$$

Note that higher values of $\Delta_{A_c}(s)$ imply greater probability that an activity in $A_c$ is assigned to $c$, given that $s$ is used as a motif. In particular, higher the value of $\Delta_{A_c}(s)$, higher will be the value of $\gamma$. Similarly, higher the value of $\sum_{c' \in C \setminus \{c\}} \Delta_{A_{c'}}(s)$, higher the value of $\lambda$.

We include a sequence $s$ as a context in the model $M_c$ iff

$$\Psi_c(s) > K \times log(\ell) \qquad (14)$$

where $\ell$ is the total length of all the activities in $A$, while $K$ is a user defined parameter. The term $K \times log(\ell)$ represents added complexity of the model $M_c$, by using $s$ as opposed to *suffix*($s$) as a context, which is shorter in length and occurs at least as often as $s$. The higher the value of $K$ the more parsimonious the model will be.

Equation 14 selects sequences that both appear regularly and have good classification and predictive power - and hence can be thought of as event-motifs. Work done in [Ron et al.1996] shows how the motifs in a *VMMC* can be compactly represented as a tree. Work done in [Apostolico and Bejerano2000] presents a linear time algorithm that constructs such a tree by first constructing a data structure called a *Suffix Tree* to represent all subsequences in the training data $A$, and then by pruning this tree to leave only the sequences representing motifs in the *VMMC* for some activity-class. We follow this general approach by using Eq 14 as our pruning criterion.

## 7 Experiments & Results

We performed experiments on extensive data-sets collected from two active environments. For both of our experimental setups, we chose the value of $n$ for the $n$-grams to be equal to 3 (*tri-grams*). This is done with the understanding that it encodes the past, present and future information of an event (essentially following second order Markov assumption).

### 7.1 Loading Dock Scenario

To test our proposed algorithms, we collected video data at the Loading Dock area of a retail bookstore. To visually span the area of activities in the loading dock, we installed two cameras with partially overlapping fields of view. A schematic diagram with sample views from the two cameras is shown in Figure 3. Daily activities from 9a.m. to 5p.m., 5 days a week, for over one month were recorded. Based on our observations of the activities taking place in that environment, we constructed an event vocabulary of 61 events. Every activity has a known starting event, i.e. *Delivery Vehicle Enters the Loading Dock* and a known ending event, i.e. *Delivery Vehicle Leaves the Loading Dock*. We used 150 of the collected instances of activities, that were manually annotated using our defined event-vocabulary of 61 events. The interaction of some low-level perceptually distinguishable primitives constitute each of these 61 events. For the Loading Dock environment, we used 10 such primitives: *Person, Cart, Delivery Vehicle(D.V.), Left Door of D.V., Right Door of D.V., Back Door of D.V., Package, Doorbell, Front Door of Building, Side Door of Building*.

### 7.2 House Scenario

To test our proposed algorithms on the activities of House environment, we deployed 16 strain gages at different locations in a house, each with a unique identification code. These transducers registered the time when the resident of the house walked over them. The data was collected daily for almost 5 months (151 days - each day is considered as an individual activity). Whenever the person passes near a transducer at a particular location, it is considered as the occurrence of a unique event. Thus our event vocabulary in this environment consists of 16 events. Figure 4 shows a schematic top-view of this environment.

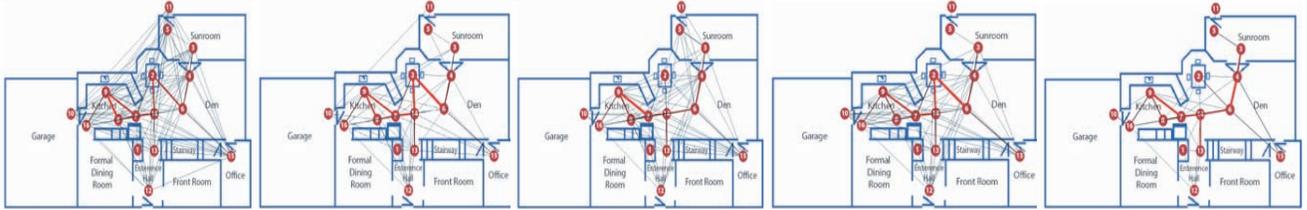

Figure 5: Visualization of the structural differences between the discovered activity-classes. Thick lines with brighter shades of red indicate higher frequency.

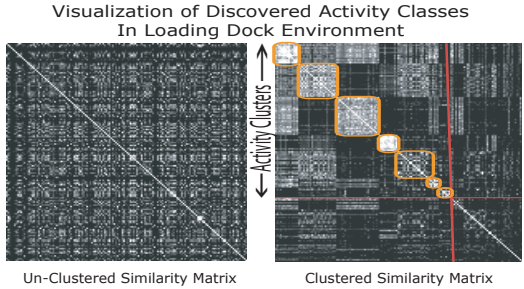

Figure 6: Each row represents the similarity of a particular activity with the entire activity training set. White implies identical similarity while black represents complete dissimilarity. The activities ordered after the red cross line in the clustered similarity matrix were dissimilar enough from all other activities as to not be included in any non-trivial maximal clique.

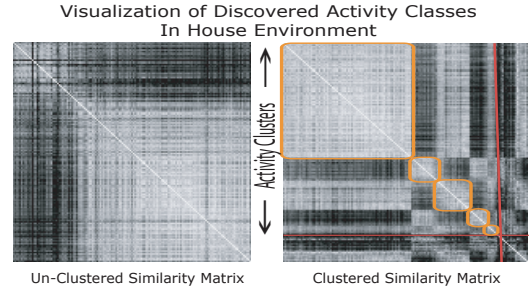

Figure 7: Visualization of similarity matrices before and after class discovery for the House Environment.

### 7.3 Discovered Activity-Classes & Motifs

**Loading Dock Scenario**

Of the 150 training activities, we found 7 classes (maximal cliques), with 106 activities as part of any one of the discovered class, while 44 activities being different enough to be not included into any non-trivial maximal clique. The visual representation for the similarity matrices of the original 150 activities and the arranged activities in 7 clusters is shown in Figure 6. These discovered activity-classes were then provided to our motif finding framework which discovered multiple motifs of various lengths, ranked by their respective bit-gains (class-characterization ability). Analysis of the discovered classes reveals a strong structural similarity amongst the class members, while the discovered motifs show ability to characterize the membership class efficiently. A brief description of the discovered activity-classes and the corresponding motifs with the maximum bit-gain is given in Figure 8.

**House Scenario**

Of the 151 activities captured over a little more than 5 months, we found 5 activity-classes (maximal cliques), with 131 activities as members of any one of the discovered class, and 20 activities being dissimilar enough not to be a part of any *non-trivial* maximal clique (see Figure 7). A brief description of the discovered activity-classes and the corresponding motifs with maximum bit-gain is given in Figure 9. To better illustrate the structural differences in the discovered activity-classes, a visualization of the normalized frequency-counts of the person's trajectory between different locations is shown in figure 5.

### 7.4 Subjective Assessment of Evaluation

The method defined above would, by construction, find activity-classes and the characterizing event-motifs. This begs the question as to how valid are the discovered activity-classes and the characterizing event-motifs. Our final goal is to design a system that would be able to discover and characterize human-interpretable activity-classes. Keeping this thought in mind, we performed a limited user test to subjectively assess the performance of our system involving 7 participant. For each participant, 2 of the 7 discovered activity classes were selected from the Loading Dock environment. Each participant was shown 6 example activities, 3 from each of the 2 selected activity-classes. The participants were then shown 6 motifs, 3 for each of the 2 classes, and were asked to associate each motif to the class that it best belonged to. Their answers agreed with our systems 83% of the time, *i.e.*, on average a participant agreed with our system on 5 out of 6 motifs. The probability of agreement on 5 out of 6 motifs by random guessing[3] is only 0.093.

---
[3]According to the binomial probability function the chance of randomly agreeing on 5 out of 6 motifs is $C_5^6(0.5)^1(0.5)^5$.

| | Class Description | Discovered Event Motif |
|---|---|---|
| Class 1 | UPS delivery-vehicles that picked up multiple packages | Person places package into back door of delivery vehicle → Person enters into side door of building → Person is empty handed → Person exists from side door of building → Person is full handed → Person places package into back door of delivery vehicle. |
| Class 2 | Delivery trucks and vans that dropped off a few packages using hand cart. | Cart is full → Person opens front door of building → Person pushes cart into front door of building → Cart is full → Person closes front door of building → Person opens front door of building → Person exists from front door of building → Person is empty handed → Person closes front door of building. |
| Class 3 | A mixture of car, van, and truck delivery vehicles that dropped off one or two packages without needing a hand cart. | DV drives in forward into LDA → Person opens left door of DV → Person exists from left door of DV → Person is empty handed → Person closes the left door of delivery vehicle. |
| Class 4 | Delivery trucks that dropped off multiple packages, using hand carts, that required multiple people. | Person opens back door of DV → Person removes package from back door of DV → Person removes package from back door of DV → Person removes package from back door of DV → Person removes package from back door of DV → Person removes package from back door of DV. |
| Class 5 | Delivery-vehicles that picked up and dropped-off multiple packages using a motorized hand cart and multiple people. | Person closes front door of building → Person removes package from cart → Person places package into back door of DV → Person removes package from cart → Person places package into back door of DV → Person removes package from cart → Person places package into back door of DV. |
| Class 6 | Delivery trucks that dropped off multiple packages using hand carts. | Person Removes Cart From Back Door of DV → Person Removes Package From Back Door of DV → Person Places Package Into Cart → Person Places Package Into Cart → Person Removes Package From Back Door of DV → Person Places Package Into Cart → Person Removes Package From Back Door of DV → Person Places Package Into Cart. |
| Class 7 | Van delivery-vehicles that dropped off one or two packages without needing a hand cart. | Person closes back door of DV → Person opens left door of DV → Person enters into left door of DV → Person is empty handed → Person closes left door of DV. |

Figure 8: The first column of the table shows the description of the discovered activity-classes, while the second column shows the highest bit-gain event-motif for each activity-class in the Loading Dock data-set. DV stands for Delivery Vehicle, while LDA stands for Loading Dock A.

| | Class Description | Discovered Event Motif |
|---|---|---|
| Class 1 | Activities lasting for the entire length of days where the person's trajectory spans the entire house space. Most of the time was spent in the area around the Kitchen and the Dining Table. | Alarm → Kitchen entrance → Fridge → Sink → Garage door (inside). |
| Class 2 | The person moves from from kitchen to the stairway more often. Further more, as opposed to cluster 1, the person does not go from the Office to the Sum Room area. | Stairway → Fridge → Sink → Cupboard → Sink. |
| Class 3 | The person spends more time in the areas of Den and the living-room. Moreover, he visits the Sun-room more often. | Stairway → Dining Table → Den → Living-room Door → Sun-room → Living-room door → Den. |
| Class 4 | The person spends most of the day in the Kitchen and the Dining Room. The duration for which she stays in the house is small for this sub-class. | Den → Living-room door → Den → Kitchen Entrance → Stairway. |
| Class 5 | The person moves from Dining Room to the Sun Room more often. The duration for which she stays in the house is significantly smaller than any other sub-class. | Fridge → Dining Table → Kitchen Entrance → Fridge → Sink |

Figure 9: The description of the discovered activity-classes in the House Data. The second column shows the discovered motifs with highest bit-gain for each class.

# 8 Discussion

The discovered activity-classes both for the Loading Dock and the House data-sets, are subjectively semantically coherent and divide their respective activity space discriminatively. The fundamental differences between various classes in the Loading Dock environment are dictated by the fact whether the activities were of deliver or pick-up, how many people were involved in the activity, how many packages were moved, and what type of delivery vehicle was used. For the House environment, these differences consist of how long does a person stay in the house, and what time of the year it is.

Figures 6 and 7 show that the activities performed in the Loading Dock environment are structurally more well defined than those performed in the House environment. This is because our vocabulary for the Loading Dock environment consists of semantically meaningful events, which can encode the underlying activity structure efficiently. For the House environment, the events are simply the locations where a person went, and are not particularly designed to encode the underlying structure of the activities.

It can be observed from table 8 and 9 that the discovered motifs of membership classes efficiently characterize these classes. Note that the discovered motifs for activity-classes where package *delivery* occurred, have events like *Person Places Package In The Back Door Of Delivery Vehicle* and *Person Pushes Cart In The Front Door of Building→ Cart is Full*. On the other hand event-motifs for activity-classes where package *pick-up* occurred, have events such as *Person Removes Package From Back-Door Of Delivery Vehicle* and *Person Places Package Into Cart*. Similarly, The motifs for the House environment capture the position where the person spends most of her time and the order in which she visits the different places in the house.

# 9 Conclusions and Future Work

In this work we introduced a novel activity representation as bags of event n-grams and posed the question of unsupervised activity discovery as a graph theoretic problem of finding maximal cliques in edge weighted graphs. We demonstrated how variable-memory Markov chains can be used to extract event-motifs that can compactly characterize activity-classes.

In the future we plan to use the discovered activity-classes and their learned characterizations as event-motifs for on-line activity classification and to detect non-regularities in activities. Finally, we intend to perform more extensive user-study to test the performance of our system.